\pdfoutput=1

\documentclass[11pt]{article}

\usepackage[final]{acl}

\usepackage{ragged2e}
\usepackage{hyperref}
\usepackage{multirow}
\usepackage{graphicx}
\usepackage{xcolor}
\usepackage{xspace}
\usepackage{color}
\usepackage{colortbl}
\usepackage{tablefootnote}
\usepackage{pifont}
\usepackage{makecell}
\usepackage[most]{tcolorbox}
\usepackage{framed}
\usepackage{mdframed}
\usepackage{subfigure}
\usepackage{caption}
\usepackage{longtable}
\usepackage{float}
\usepackage{booktabs}
\usepackage{arydshln} 

\newcommand\Matrix{\mathbf} 
\newcommand\Tensor{\mathcal}
\usepackage{amsfonts}
\usepackage{pgfplots}
\usepackage{pgfplotstable}
\usepackage{booktabs}

\newcommand{\paratitle}[1]{\vspace{1.5ex}\noindent\textbf{#1}}
\newcolumntype{H}{>{\setbox0=\hbox\bgroup}c<{\egroup}@{}}

\newcommand{\ie}{\emph{i.e.,}\xspace}
\newcommand{\vs}{\emph{vs.}\xspace}

\newcommand{\eg}{\emph{e.g.,}\xspace}

\newcommand{\ignore}[1]{}

\usepackage{algorithm}
\usepackage{algpseudocode}
\usepackage{amsthm}
\usepackage{amsmath}
\usepackage{tikz}

\usepackage{amsmath}

\usepackage{times}
\usepackage{latexsym}

\usepackage[T1]{fontenc}

\usepackage[utf8]{inputenc}

\usepackage{microtype}

\usepackage{inconsolata}

\title{Unlocking Data-free Low-bit Quantization with Matrix Decomposition for KV Cache Compression}

\author{
	Peiyu Liu$^{1,2}$\thanks{$\ $ This work was done during an internship at Huawei.} ,
	Ze-Feng Gao$^{2,4}$\ ,
	Wayne Xin Zhao$^{2}$\thanks{$\ $ Corresponding author.}\ ,
	\\ \textbf{Yipeng Ma$^{3}$\ ,}\textbf{Tao Wang$^{3}$\ ,}
	\textbf{Ji-Rong Wen}$^{2,6}$\\
        $^1$ School of Information Technology and Management, University of International Business and Economics \\
	$^2$ Gaoling School of Artificial Intelligence, Renmin University of China, $^3 $Huawei Technologies Co., Ltd.\\
	$^4$ Department of Physics, $^5$ School of Information, Renmin University of China\\
	\tt{liupeiyustu@163.com,\{zfgao,jrwen\}@ruc.edu.cn}, \\\tt{batmanfly@gmail.com,\{mayipeng,wangtao10\}@huawei.com}
}
\begin{document}
\maketitle
\begin{abstract}
Key-value~(KV) caching is an important technique to accelerate the inference of large language models~(LLMs), but incurs significant memory overhead.  
To compress the size of KV cache, existing methods often compromise precision or require extra data for calibration, limiting their practicality in  LLM deployment. 
In this paper, we introduce \textbf{DecoQuant}, a novel data-free low-bit quantization technique based on tensor decomposition methods, to effectively compress KV cache. 
Our core idea is to adjust the outlier distribution of the original matrix by performing tensor decomposition, so that the quantization difficulties are migrated from the matrix to decomposed local tensors. Specially, we find that outliers mainly concentrate on small local tensors, while large tensors tend to have a narrower value range. Based on this finding, we propose to apply low-bit quantization to the large tensor, while maintaining high-precision representation for the small tensor. 
Furthermore, we utilize the proposed quantization method to compress the KV cache of LLMs to accelerate the inference and develop an efficient dequantization kernel tailored specifically for DecoQuant.
Through extensive experiments, DecoQuant demonstrates remarkable efficiency gains, showcasing up to a $\sim$75\% reduction in memory footprint while maintaining comparable generation quality. 
\end{abstract}
\section{Introduction}
Large language models~(LLMs)~\cite{touvron2023llama,wayne2023survey} have made significant strides in advancing the progress of language intelligence.  
However, these large-sized models often 
incur higher inference latency,  
bringing  significant challenges to practical deployment. 
Therefore, it is urgent to reduce the running overhead of LLMs. 

To optimize the efficiency of LLMs during the inference process, a commonly used technique is \emph{key-value~(KV) caching}~\cite{pope2022efficiently}. 
In implementation, KV caching involves the storage of historical tokens associated with the attention key and value tensors of
each layer, offering accelerated inference by trading increased memory consumption for a reduction in redundant calculations.
However, applications of long-content generation, such as story generation and long demonstrations for in-context learning tasks, would lead to a significant increase in the size of the KV cache, resulting in unaffordable storage costs~\cite{zhang2023h2o,liu2023scissorhands}. 
In addition, managing a large cache often involves frequent I/O read and write operations, leading to considerable latency. 
The issue becomes even more severe when I/O operations need to span across multiple machines~\cite{patel2023splitwise}. 
Therefore, we need to compress KV cache of large models to optimize the inference process.

Considering the above issues, considerable efforts have concentrated on KV cache compression to enhance inference efficiency. As a typical approach, recent work~\cite{zhang2023h2o,mu2023learning-gist} prunes tokens to keep the KV cache within a small size. This approach, while alleviating memory overhead, potentially 
leads to information loss in long text generation.
Furthermore, although post-quantization methods preserve all preceding text, low-bit quantization often results in substantial model performance degradation. This is primarily attributed to the common challenge of outlier problems in activation value quantization~\cite{tim2022llmint8}. Additionally, current quantization techniques still rely on calibration or training~\cite{elias2022gptq,xiao2023smoothquant} to retain the model performance, thus imposing practical limitations in data-constrained settings (\eg privacy data). 
This further highlights the need for a data-free approach to KV cache compression.

To effectively quantize the KV cache~(essentially activation values), we draw inspiration from SmoothQuant~\cite{xiao2023smoothquant}, which suggests that the issue of outliers can be transferred across multiple modules, by migrating the quantization difficulty to weights. However, unlike SmoothQuant, 
we take an improved approach by directly migrating the quantization difficulty by performing matrix decomposition on the activation values themselves, without comprising the precision of the weights.
The underlying principle is that matrix decomposition can potentially adjust the outlier distribution of the original matrix~\cite{liu2021enabling}, so that the decomposed local tensors or matrices are easier to quantize. 

To this end, in this paper, we propose an effective matrix \emph{\underline{Deco}}mposition based \emph{\underline{Quant}}ization method namely \textbf{DecoQuant}, to alleviate the quantization error due to outliers.
Our approach is developed based on an important empirical finding: when performing tensor decomposition (\ie  Matrix Product Operator), the value range of the large local tensor (consisting of the major proportion of parameters) becomes narrower, indicating fewer outliers to be resolved in quantization. 
Based on this finding, we propose a local tensor based quantization method, in which we apply low-bit quantization to the large tensor, while maintaining high-precision representation for the small tensor. In this way, we can achieve a lower quantization error when reconstructing the original matrix by multiplying all the local tensors.
Furthermore, we utilize the proposed quantization method to compress the KV cache of LLMs to accelerate the
 inference rate, and further develop an  efficient dequantization kernel tailored specifically for DecoQuant.


DecoQuant provides an effective quantization approach for LLMs, which can  compress KV cache to accelerate the inference rate.  
It is featured by two major merits, namely (1)~\emph{fully data-free} by  eliminating the need for complex calibration mechanisms and (2)~\emph{highly flexible} by supporting the quantization for weights only, activations only  as well as both simultaneously. 
Extensive experiments have demonstrated the effectiveness of the proposed approach in reducing the memory consumption of the KV cache and achieving competitive performance. With nearly lossless performance, we can achieve 4-bit KV cache quantization and 8-bit quantization for both weights and activations.
\section{Preliminary}
In this section, we present the  background for our approach about LLM inference and quantization. 

\paratitle{LLM Inference and KV Caching}. Typically, LLMs generate the next token in a two-step  process~\cite{wayne2023survey,zhong2024distserve}:~(1) \emph{prefilling} phase, in which LLMs generate the first token based on the prompt, and (2) \emph{decoding} phase, in which the rest tokens are generated one by one in an auto-regressive manner. Specifically, the decoding phase dominates the inference latency in long-text generation~(\eg story writing). 
A common practice to accelerate the decoding phase is key-value~(KV) caching~\cite{pope2022efficiently}, which stores previously seen tokens to avoid recomputing of attention key and value tensors.  However, the size of the KV cache increases linearly with the generation length which poses a memory-bounded challenge. 
Furthermore, the increase in computing power has increased substantially~(\eg 3.4x from A100 to H100) while the communication improvements have lagged behind~(\eg only 1.6x from A100 to H100). This highlights the vital need to address memory compression for the KV cache.

\paratitle{LLM Quantization.} Quantization maps a floating-point number into low-bit integers, which can largely reduce the model size and inference costs of LLMs~\cite{lin2023awq,elias2022gptq,tim2022llmint8}.  
We follow~\citet{xiao2023smoothquant} and 
use symmetric quantization for simplicity
while the discussion for asymmetric cases is similar by adding a zero-point~\cite{jacob2018quantization}. 
Generally speaking, there are two major kinds of matrices to be quantized in LLMs, namely \emph{weights} and \emph{activations}. In the context of quantizing LLMs, there are typically two approaches: quantizing only the weights to preserve model accuracy or quantizing both the weights and activation values to enhance the hardware compatibility. 
Formally, the quantization process of a single matrix can be expressed as the following formula:
\begin{equation}
    \hat{\Matrix{W}}=\left \lceil \frac{\Matrix{W}}{\Delta} \right \rfloor, \Delta=\frac{max(\left | \Matrix{W} \right | )}{2^{(m-1)}-1},
    \label{equ-mpo}
\end{equation}
where $\Matrix{W}$ is the floating-point matrix, $\hat{\Matrix{W}}$ is the quantized conterpart, and $\Delta$ is the quantization step size, $\left \lceil \cdot \right \rfloor$ is the rounding function and $m$ is the number of bits. 
However, it is practically difficult to set a suitable value for $\Delta$,  mainly due to the existence  of \emph{outliers} (those significantly deviate from the majority of values)~\cite{tim2022llmint8}.
Therefore, we aim to mitigate the impact of outliers to achieve the quantization compression of the KV cache.

\paratitle{Tensor Decomposition}. 
Tensor decomposition~\cite{rabanser2017introduction,kolda2009tensor} is a standard algorithm to factorize a matrix into a sequential product of local tensors. 
Specially, we adopt Matrix Product Operator~(MPO)~\cite{liu2021enabling} as the decomposition strategy.
Formally we describe the process of decomposing a matrix $\Matrix{W}\in\mathbb{R}^{I\times J} $ using MPO as follows:
\begin{equation}
    \textsc{MPO}~(\Matrix{W})=\prod_{k=1}^{n} \Tensor{T}_{(k)}[d_{k-1},i_k,j_k,d_k],
    \label{eq:mpo}
\end{equation}
where $\Tensor{T}$ denotes the local tensor with size $d_{k-1}
\times i_k\times j_k\times d_k$ in which $\prod_{k=1}^{n}i_k=I, \prod_{k=1}^{n}j_k=J$ and $n$ represents the number of local tensors. 
We refer to the decomposed tensors as local tensors. When $n=2$, we designate the tensor with a larger parameter count as $\Tensor{T}_L$~(\ie the \emph{central tensor} in \citealp{liu2021enabling}), and the one with fewer parameters as $\Tensor{T}_S$. 
With MPO decomposition, we can reorganize and aggregate information within specific tensors providing us with the opportunity to effectively distinguish outliers.



\section{Methods}
In this section, we present an effective matrix \emph{\underline{Deco}}mposition based \emph{\underline{Quant}}ization method namely \textbf{DecoQuant},  to alleviate the quantization error due to outliers. We further utilize this method to quantize the KV cache for efficient inference of LLMs. 
\subsection{DecoQuant: Matrix Quantization based on Decomposition}
\label{sec-tensor_dequant}
Basically, our approach aims to employ tensor decomposition to adjust the outlier distribution in the original matrix, so as to mitigate the quantization difficulty. As will be introduced, decomposed local tensors tend to exhibit fewer outliers within their value distributions, indicating a potential opportunity for improving  quantization accuracy. In what follows, we first study the distribution of outliers in local tensors and then propose an effective  quantization approach based on tensor decomposition.   

\begin{figure}[t]
\centering
\subfigure[Analysis for $\Tensor{T}_L$.]{
\begin{minipage}[ht]{0.23\textwidth}
\centering
\includegraphics[width=1.1\textwidth]{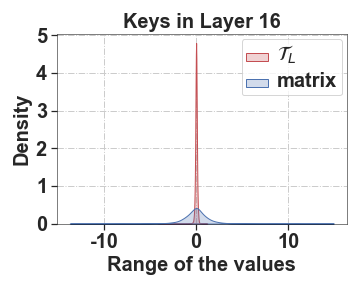} 
\label{fig-a}
\end{minipage}
}
\subfigure[Analysis for $\Tensor{T}_S$.]{
\begin{minipage}[ht]{0.23\textwidth}
\centering
\includegraphics[width=1.1\textwidth]{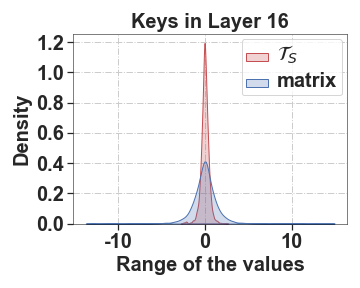} 
\label{fig-b}
\end{minipage}
}
\caption{Outlier distributions of local tensors and matrices. ``Keys'' are extracted from the output features of value projections in the 16th layer of LLaMA-7B.
Investigations of other structures can refer to Appendix~\ref{app-outlier}.}
\label{fig-analysis}
\end{figure}

\paratitle{Outlier Distributions in Local Tensors.}
We are mainly concerned with the KV cache matrices, as they highly affect the inference latency~\cite{zhang2023h2o,patel2023splitwise,liu2023scissorhands}.  
Without loss of generality, we consider $n$=2 for MPO decomposition and take the key state matrix, \ie $\Matrix{K}$, as  example:
\begin{equation}
    \textsc{MPO}(\Matrix{K})=\Tensor{T}_L \times \Tensor{T}_S.
\end{equation}
A property of MPO is that it can adjust the distribution of parameters~(\ie $\{d_{k-1},i_k\,j_k\,d_k\}$ in Equation~\ref{equ-mpo}) in these local tensors. Specially, we take a biased decomposition, where $\Tensor{T}_L$ takes a large proportion of parameters (\ie 99.4\%) while $\Tensor{T}_S$ only takes a small proportion of parameters  (\ie 0.6\%). Such a large tensor $\Tensor{T}_L$ is also called \emph{central tensor}~\citep{liu2021enabling}, since it contains the large body of information of the original matrix. 
Further, we examine the change of the outlier distribution in both tensors.  
In Figure~\ref{fig-analysis}, we can observe an interesting finding that the value distribution of the large tensor $\Tensor{T}_L$ becomes much narrower than the original matrix and the small tensor $\Tensor{T}_S$. In other words, it becomes easier to quantize $\Tensor{T}_L$ with fewer bits due to the limited value distribution. Despite that it is still difficult to quantize $\Tensor{T}_S$, it is noted that $\Tensor{T}_S$ only contains a small number of parameters, and we can apply higher quantization precision with an overall small cost.  


\paratitle{Local Tensor Quantization.}
Based on the above discussion, we introduce a novel data-free quantization method based on matrix decomposition. 
The key idea of our method is that through tensor decomposition, the quantization difficulties~(\ie outliers) can be transferred from the original matrix to its \emph{small  local tensors}. 
Thus, we can consider applying \emph{low-precision quantization} to the large tensors, while maintaining \emph{high-precision representation} for the small tensors. In this way, we can achieve a lower quantization error when reconstructing the original matrix.  
Specially, our approach involves a two-step quantization process which is shown in Figure~\ref{fig:dequant}: 
~(1) First, we utilize MPO to factorize the original matrix into two higher-dimensional local tensors~(\ie $\Tensor{T}_S$ and $\Tensor{T}_L$).
As shown in Figure~\ref{fig-analysis}, an important characteristic is that $\Tensor{T}_L$, which occupies a significant portion of the parameters, has a much smaller distribution of outliers than that of the original matrix. 
~(2) Thus, at the second step, we focus on quantizing the larger tensor $\Tensor{T}_L$ into $B$-bit integers ($B<16$) while preserving 16-bit precision for $\Tensor{T}_S$ to achieve a lower quantization error (with verified effectiveness in Section~\ref{sec-detailanalyais}). 

\subsection{Efficient Inference based on DecoQuant}
\begin{figure}
    \centering
    \includegraphics[width=0.3\textwidth]{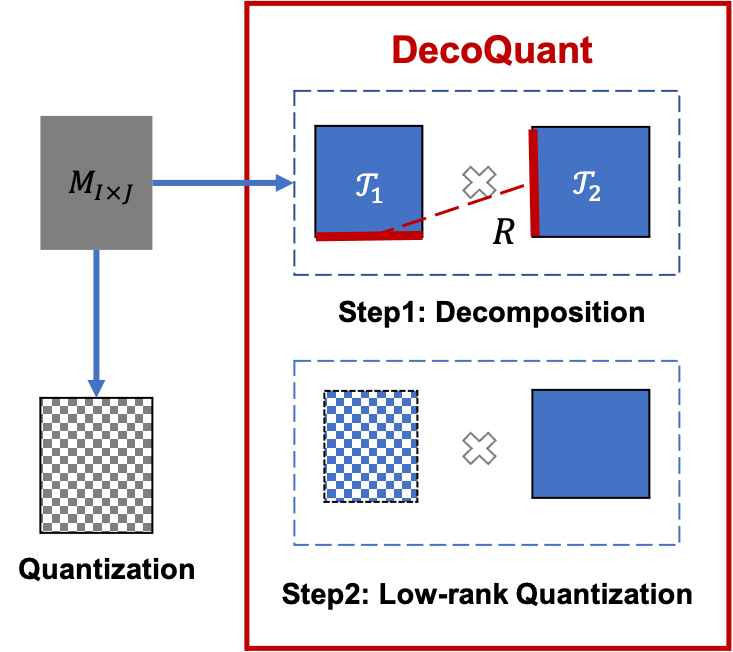}
    \caption{Matrix quantization based on DecoQuant. The alternating black/white and blue/white squares in the figure denote quantized matrices.}
    \label{fig:main}
\end{figure}
Building upon the DecoQuant approach discussed in Section~\ref{sec-tensor_dequant}, which achieves data-free matrix quantization through the quantization of decomposed local tensors, our primary objective is to compress the KV cache of LLMs to accelerate the inference rate. 
The key idea is to quantize the KV cache into a low-bit representation while preserving FP16 precision during computation. Additionally, we have developed a consolidated and efficient dequantization kernel tailored specifically for DecoQuant.


\paratitle{KV Cache Quantization.}
To introduce our method, we consider a typical $L$-layer Transformer model with $D$ dimensions, where the input text consists of $T$ prompt tokens. Then we consider compressing \textit{key} and \textit{value} cache for two phases of LLM inference separately.~(1) Prefilling phase:  The key and value cache are initially obtained after the generation of the first token, \ie $\Matrix{K},\Matrix{V}\in \mathbb{R}^{T\times D}$. Given the relatively large size of the matrices, 
we utilize the DecoQuant technique offline on the KV cache to alleviate the computational overhead induced by decomposition.~(2) Decoding phase: 
The size of the KV cache grows linearly with the sequence length,
\ie $\Delta\Matrix{K},\Delta\Matrix{V}\in \mathbb{R}^{1\times D}$. To alleviate the increased computational workload due to frequent quantization, we perform DecoQuant only when the cache accumulates 
a certain length~(\eg 1$k$).
In particular,  DecoQuant supports quantization for weights only~(WxA16), activations only~(W16Ax), as well as both simultaneously~(WxAx),  significantly expanding its applicability. 
Next, we will describe the dequantization process when the key and value cache are recovered to FP16 precision for computation. 

\paratitle{Kernel Fusion for Dequantization.}
\begin{figure}[t]
    \centering
    \includegraphics[width=0.4\textwidth]{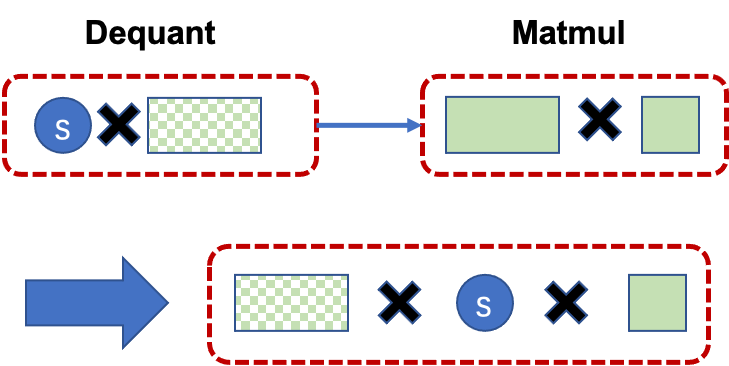}
    \caption{Operator fusion for dequantization.}
    \label{fig:dequant}
\end{figure}
Kernel fusion~\cite{wang2010kernel} is a technique that combines multiple separate computational kernels into a \emph{single, more efficient} kernel. Essentially, it allows multiple kernels to be executed as a whole unit and thus reduces the overhead and latency in processing. 
In our approach, the dequantization of DecoQuant involves operations that convert integers to floating-point values~(\ie dequantization operator of quantization scales and integer values) and that reconstruct local tensors to matrices~(GeMM operator of $\Tensor{T}_L$ and $\Tensor{T}_S$). These two operations may involve an additional data movement overhead between GPU compute units and the main memory which leads to increased latency.
To address this issue, we design specific kernel fusion methods for 2/4/8-bit values by fusing the dequantization operator with the next GeMM operator~(as shown in Figure~\ref{fig:dequant}), which streamlines the execution pipeline and improves computational efficiency.
By doing this, we can effectively alleviate the computational delay caused by data-movement overhead~(see Section~\ref{sec-analysis-complexity} for specific experiments).
\ignore{
\begin{algorithm}[t]
\small
    \caption{KV cache quantization.\textcolor{blue}{TODO}} 
    \begin{algorithmic}[1]
    \Require \textit{prompt}: The prompt or demonstrations before the input; \textit{inputs}: The input to LLMs; \textit{bit}: The precision level of quantization; \textit{L}: The total layers of the LLM.
    \State \textbf{Prefilling phase}
    \State $\Matrix{K}, \Matrix{V} \gets$ model(prompt)
    \State $\hat{\Matrix{K}}, \hat{\Matrix{V}} \gets$ Quantize($\Matrix{K}, \Matrix{V}$, bit)
    \State \textbf{Decoding phase}
    \While {not finish}
        \State hiddens $\gets$ Emb(\textit{inputs})
        \For {$i = 1 \gets L$}
            \State $\Matrix{K}^{(i)},\Matrix{V}^{(i)} \gets $ Dequantize($\hat{\Matrix{K}}^{(i)}, \hat{\Matrix{V}}^{(i)}$)
            \State hiddens $\gets$ Layer$_i$(hiddens)
        \EndFor
    \State next token $\gets$ lm\_head(hiddens)
    \EndWhile
    \State \Return outputs
\end{algorithmic}
\label{algocomplete}
\end{algorithm}
}

\subsection{Discussion} 
In this part, we present the overhead analysis of the proposed approach and then compare it with existing work. 

\paratitle{Compression Ratio and Time Complexity.}
In this part, we assess the memory compression ratio and time complexity of DecoQuant. While the tensor parameters obtained through DecoQuant are slightly larger than the original matrix, 
the significant storage reduction primarily stems from converting the majority of $\Tensor{T}_L$ parameters from FP16 to $B$-bit integers, allowing for a more efficient representation of the tensor and a decrease in storage requirements.
This reduction is quantified as the compression ratio~($\mu$), which is calculated as:
\begin{equation}
    \mu=\frac{\#(\Tensor{T}_L)\times B+\#(\Tensor{T}_S)\times 16+\#(\Delta)}{\#(\Matrix{W})\times 16},
\end{equation}
where $\#(\cdot)$ denotes the count of values. 
Due to the significantly smaller number of parameters in $\Tensor{T}_S$ and $\Delta$ compared to $\Tensor{T}_L$, the compression ratio typically approximates $B/16$. 
For inference time, DecoQuant significantly reduces communication costs with 4-bit KV cache. This results in a speedup of ~1.25x under conditions of generating an output of 6k tokens~(Section~\ref{sec-analysis-complexity}).

\paratitle{Comparison with Existing Work.} 
We compare our method with existing methods~(including RTN, LLM.int8(), SmoothQuant, GPTQ, and AWQ) from the perspectives of quantization settings and requirement for extra data, with results presented in Table~\ref{tab-advantage}. We find that, similar to RTN, our method can support all quantization settings, including weight only~(WxA16), activation only~(W16Ax), and simultaneous~(WxAx) in a data-free style. In contrast, other methods typically only support a subset of these settings~(such as GPTQ and AWQ supporting only WxA16, while LLM.int8() supports WxAx) thereby limiting their practical application. Additionally, some methods require extra data for calibration. However, obtaining calibration data for scenarios involving sensitive user privacy can be challenging. Thus we primarily focus on RTN as our comparative baseline, introducing other methods only as needed.

\begin{table}[t]
    \centering
    \small
    \caption{DecoQuant facilitates data-free quantization for weights only~(WxA16), activations only~(W16Ax), as well as both simultaneously~(WxAx). RTN denotes the vanilla round-to-nearest quantization~\cite{lin2023awq}.}
    \begin{tabular}{cccccc}
        \toprule
        Methods             & \multicolumn{3}{c}{Support} & \multirow{2}{*}{Data-free}\\ 
                            & WxA16     & W16Ax   & WxAx  & \\ \midrule
        RTN                 & \ding{52} & \ding{52} & \ding{52} & \ding{52} \\
        GPTQ                & \ding{52} & \ding{56} & \ding{56} & \ding{56} \\
        AWQ                 & \ding{52} & \ding{56} & \ding{56} & \ding{56} \\
        LLM.int8()          & \ding{56} & \ding{56} & \ding{52} & \ding{52} \\
        SmoothQuant         & \ding{56} & \ding{56} & \ding{52} & \ding{56} \\ \midrule
        \textbf{DecoQuant}  & \ding{52} & \ding{52} & \ding{52} & \ding{52} \\ \bottomrule
    \end{tabular}
    \label{tab-advantage}
\end{table}

\section{Experiments}
We mainly evaluate the DecoQuant on the language modeling task to compare it with other quantization approaches. Then we explore its zero-shot generalization ability in open-ended document generation. Finally, we quantitatively measure the effect of KV cache compression on system throughput.
\subsection{Experimental Setup}
\paratitle{Datasets and Implementation.} For language modeling tasks, we conduct our experiments on LAMBADA~\cite{paperno2016lambada} dataset, which is a widely used dataset evaluating the ability of language models to capture long-range dependencies and contextual understanding in text. To evaluate the effectiveness of DecoQaunt in downstream tasks, we follow~\citet{chevalier2023adapting} and consider five tasks~(AG News, Subj, MR, Boolq and RTE) for in-context learning setting. The accuracy is reported to measure the quality of the next token prediction task of different models as well as the downstream tasks.
We consider popular large language models with various sizes including LLaMA~(7B and 13B)~\cite{touvron2023llama} and OPT~(1.3B and 6.7B)~\cite{zhang2022opt}. For the quantization setting, we follow~\cite{xiao2023smoothquant} and quantize the weights, activations and KV cache into different bit-precisions~(2/4/8/16 bits). The code to reproduce the results of this paper can be found at~\url{https://github.com/lpyhdzx/DecoQuant_code}.




\paratitle{Baselines.} We introduce popular baseline quantization methods for KV cache compression.

$\bullet$ Round-to-nearest~(RTN,~\citealt{lin2023awq}). RTN maps a real value to an integer value through a naive rounding operation. 

$\bullet$ SmoothQuant~\cite{xiao2023smoothquant}. SmoothQuant smooths the activation outliers to weights and only supports WxAx quantization.

Some widely used quantization methods, such as GPTQ and LLM.int8(), are not considered because they cannot quantize the output activation values, thus making them unsuitable for quantization in the KV cache.

\subsection{Main Results}
\begin{table*}[ht]
    \centering
    \small
    \caption{Results when key and value modules are quantized to different levels~(denoted as W-A-). 
    ``*'' indicates the quantization results based on the calibration dataset generated using the official code.
    }
    \begin{tabular}{ccccccccccc}
        \toprule
        Setting              & Exp             & \#Bits    & Size$_{\rm{(MB)}}$ & LLaMA-7B & LLaMA-13B & OPT-1.3B & OPT-6.7B                        & Average\\ \midrule
        & FP16        & 16-16    & 46.7    & 87.8          & 89.3            & 75.4         & 81.2         & 83.4\\ \cdashline{2-8}
        & RTN         & 16-8  & 23.3    & 88.6          & 89.3            & 75.3         & 81.2         & 83.6\\ 
        & DecoQuant   & 16-8  & 23.3    & 88.6          & \textbf{89.4}   & \textbf{75.4}& 81.2         &\textbf{83.7}\\\cdashline{2-8}
        activations & RTN         & 16-4  & 11.7    & 86.0      & 88.1            & 71.7         & 80.6         & 81.6\\ 
        only    & DecoQuant   & 16-4  & 11.7    & \textbf{88.1} & \textbf{88.9}   & \textbf{73.6}& \textbf{80.9}& \textbf{82.9}    \\\cdashline{2-8}
        & RTN         & 16-2  & 5.8     & 1.0          & 0.0            & 3.5         & \textbf{4.7}& 2.3    \\ 
        & DecoQuant   & 16-2  & 5.8     & \textbf{47.1} & \textbf{58.2}   & \textbf{8.6}& \textbf{28.8}& \textbf{35.9}         \\\midrule
        & RTN         & 8-8    & 23.3    & 88.5          & 89.3            & \textbf{75.4}& 81.3         & 83.6         \\ 
        & SmoothQuant & 8-8    & 23.3    & 88.5$^*$      & 89.3$^*$                & 75.3         & 81.3         & /         \\
        & DecoQuant   & 8-8    & 23.3    & 88.5          & \textbf{89.4}   & \textbf{75.4}& 81.3         & \textbf{83.7}\\ \cdashline{2-8}
        weights & RTN         & 4-4    & 11.7    & 86.4          & 88.0            & 69.4         & 78.5         & 80.6\\ 
        \&      & SmoothQuant & 4-4    & 11.7    & 86.4$^*$          & 88.0$^*$                & 69.0         & 77.7         & /\\ 
        activations& DecoQuant   & 4-4    & 11.7    & \textbf{88.4} & \textbf{88.5}   & \textbf{70.8}& \textbf{79.1}& \textbf{81.7}         \\\cdashline{2-8}
        & RTN         & 2-2    & 5.8     & 0.0          & 0.0            & 3.6         & \textbf{3.2}& 2.0\\ 
        & SmoothQuant & 2-2    & 5.8     & 0.4$^*$              & 0.0$^*$                & \textbf{3.8}& 3.0         & /         \\ 
        & DecoQuant   & 2-2    & 5.8     & \textbf{1.3} & \textbf{3.0}   & 1.8         & 2.9         & 2.0\\\bottomrule
    
    \end{tabular}
    \label{tab-main}
\end{table*}
\paratitle{Comparison with Other Quantization Methods.}
The results on LAMBADA are shown in Table~\ref{tab-main}. Compared with FP16, all quantization methods reduce the sizes of the KV cache significantly due to low bit-precisions. Overall, we observe that DecoQuant achieves better average scores than other methods. 
We note that RTN sometimes gives better results~(LLaMA-13B), but this performance is not stable, and in other cases, it is not good. We suspect that it is related to the distribution of outliers in the model, an observation that is very similar to~\cite{tim2022llmint8}, which mentions that there is a clear difference in the distribution of outliers for large models.
When comparing different quantization settings, we find that 4-bit quantization often exhibits close performance to 16-bit performance while 2-bit models get much worse.
Interestingly, even in 2-bit quantization, DecoQuant still has a significant advantage over other methods, an observation that opens up the possibility of a 2-bit KV cache in the future, an exploration we leave to be completed in subsequent work.

\paratitle{Evaluation on Long-text Tasks.}
We evaluate DecoQuant's in-context learning capabilities using OPT models on five distinct datasets. For each dataset, we conduct experiments with varying numbers of demonstrations to investigate the impact of KV cache quantization on the contextual length. The summarized results are presented in Table~\ref{tab-icl}.
Our findings indicate that a larger number of demonstrations often results in performance improvements, as evidenced by the performance comparison, \eg 72.8 compared to 66.8 for FP16. This observation underscores the effectiveness of augmenting the contextual information.
However, when comparing the performance of RTN and DecoQuant, we observe that, on average, RTN lags behind DecoQuant. An interesting aspect of this comparison is that RTN's performance is comparable to DecoQuant's in the case of shorter contexts~(2-shot), but it notably deteriorates for longer contexts~(10-shot). This outcome reinforces the efficacy of our approach, which effectively compresses the prompt while preserving critical information.
\begin{table*}[htbp]
    \centering
    \small
    \caption{Results of in-context learning with different lengths of demonstrations.}
    \begin{tabular}{cccccccccc}
        \toprule
        Models & Exp  & \#Bits    & ICL       & Ag\_news      & Subj          & Mr            & Boolq         & RTE                      & Average\\ \midrule
        \multirow{11}{*}{OPT-1.3B} & FP16        & 16        & 0-shot    & 58.0          & 62.9          & 79.5          & 60.5          & 52.7          & 62.7\\ 
        & FP16        & 16        & 2-shot    & 64.2          & 55.1          & 86.1          & 56.9          & 45.1          & 61.5\\ 
        & FP16        & 16        & 10-shot   & 70.0          & 64.4          & 84.0          & 64.7          & 50.2          & 66.7\\ \cmidrule{2-10}
        & RTN         & 4         & 2-shot    & 61.7          & \textbf{63.1} & 81.1          & 41.1          & 45.1          & 58.4\\ 
        & DecoQuant   & 4         & 2-shot    & \textbf{62.4} & 55.8          & \textbf{87.0} & \textbf{52.2} & \textbf{46.9} & \textbf{60.9}\\\cdashline{2-10}
        & RTN         & 4         & 10-shot   & \textbf{63.6} & 51.7          & 83.7          & 63.0          & 48.7          & 62.1\\ 
        & DecoQuant   & 4         & 10-shot   & 62.6          & \textbf{69.7} & \textbf{85.6} & 63.0          & 48.4          & \textbf{65.9}\\\cdashline{2-10}
        & RTN         & 2         & 2-shot    & 33.0          & 51.7          & \textbf{55.0} & 41.8          & 53.1          & 46.9\\ 
        & DecoQuant   & 2         & 2-shot    & \textbf{40.4} & \textbf{56.0} & 52.1          & \textbf{49.7} & 52.3          & \textbf{51.6}\\\cdashline{2-10}
        & RTN         & 2         & 10-shot   & 37.6          & 53.7          & 52.7          & 39.0          & 49.5          & 51.4\\ 
        & DecoQuant   & 2         & 10-shot   & \textbf{42.4} & \textbf{65.6} & \textbf{54.1} & \textbf{43.4} & 52.7          & \textbf{66.2}\\\midrule
        \multirow{11}{*}{OPT-6.7B} & FP16        & 16        & 0-shot    & 70.9          & 61.4          & 64.3          & 63.5          & 60.3          & 64.1\\ 
        & FP16        & 16        & 2-shot    & 71.0          & 74.0          & 89.9          & 65.7          & 54.2          & 71.0\\ 
        & FP16        & 16        & 10-shot   & 53.3          & 89.8          & 86.8          & 65.7          & 57.0          & 70.5\\ \cmidrule{2-10}
        & RTN         & 4         & 2-shot    & 68.7          & 66.1          & 81.1          & 67.5          & \textbf{53.8} & 69.2\\ 
        & DecoQuant   & 4         & 2-shot    & \textbf{71.6} & \textbf{73.6} & \textbf{87.0} & \textbf{68.2} & 53.4          & \textbf{71.2}\\\cdashline{2-10}
        & RTN         & 4         & 10-shot   & 53.1          & 76.8          & 83.7          & 64.7          & \textbf{54.5} & 67.2\\ 
        & DecoQuant   & 4         & 10-shot   & \textbf{54.6} & \textbf{92.4} & \textbf{85.6} & 62.6          & 51.3          & \textbf{70.0}\\\cdashline{2-10}
        & RTN         & 2         & 2-shot    & 29.3          & 51.7          & \textbf{55.0} & 38.0          & 52.0          & 44.2\\ 
        & DecoQuant   & 2         & 2-shot    & \textbf{32.0} & \textbf{55.1} & 52.1          & \textbf{61.5} & \textbf{53.4} & \textbf{53.1}\\\cdashline{2-10}
        & RTN         & 2         & 10-shot   & \textbf{45.7} & \textbf{51.7} & 52.7          & 47.5          & 50.2          & 49.0\\ 
        & DecoQuant   & 2         & 10-shot   & 44.9          & 48.8          & \textbf{54.1} & \textbf{60.1} & \textbf{53.4} & \textbf{51.8}\\\bottomrule
    \end{tabular}
    \label{tab-icl}
\end{table*}
\subsection{Detailed Analysis}
\label{sec-detailanalyais}
\paratitle{Effectiveness of Tensor Quantization.}
\begin{figure}[t]
\centering
\subfigure[Quantization strategy.]{
\label{fig-quant_error}
\begin{minipage}[ht]{0.23\textwidth}
\centering
\includegraphics[width=1.05\textwidth]{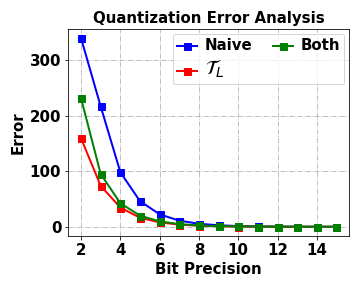} 
\end{minipage}
}
\subfigure[Length of decomposition.]{
\label{fig-length}
\begin{minipage}[ht]{0.23\textwidth}
\centering
\includegraphics[width=1.05\textwidth]{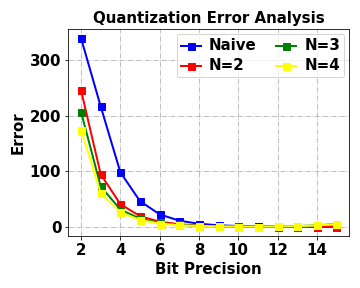} 
\end{minipage}
}
\caption{Quantization error analysis about quantization strategy and length of decomposition.}
\label{fig-effective}
\end{figure}
First, We compare the errors after quantization with different local tensors to illustrate the effectiveness of DecoQuant in mitigating the influence of outliers.
Specifically, we evaluate two variants that quantize different local tensors:~(1) quantizing large local tensors only, \ie $\Tensor{T}_L$, and~(2) quantizing both local tensors. 
The reconstruction error is shown in Figure~\ref{fig-quant_error}. We find that quantizing only the largest one~(\ie the red line) has the lowest error, followed by quantizing both~(the green line). The quantization against the matrix~(the blue one) has the largest quantization error. This demonstrates that the issue of quantization error for activations can be considerably mitigated by substituting matrix quantization with local tensor quantization. 

\paratitle{Analysis of Length of Local Tensors.}
We vary the MPO decomposition length~($n$) to assess its impact on quantization. 
Specifically, we choose $n=2,3,4$ and the results are shown in Figure~\ref{fig-length}. 
This result shows that we can further enhance the quantization by extending the length of decompositions, which validates that the tensor decomposition process is indeed beneficial in mitigating the effect of outliers on quantization.
However, the gains diminish as $n$ increases. Notably, the improvement from $n=2$ to $n=3$ is higher than from $n=3$ to $n=4$. Considering effectiveness and efficiency, we select $n=2$ for our experiments but recommend higher $n$ for higher accuracy.

\paratitle{Comparison with Other Decomposition Methods.}
\begin{figure}[t]
\centering
\subfigure[Compression ratio.]{
\label{fig-type}
\begin{minipage}[ht]{0.23\textwidth}
\centering
\includegraphics[width=1.1\textwidth]{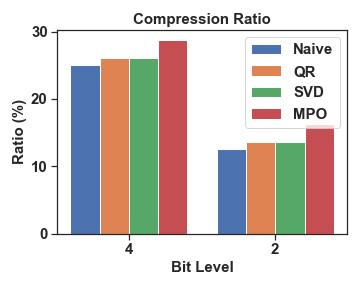} 
\end{minipage}
}
\subfigure[Quantization error.]{
\label{fig-compare_decom_error}
\begin{minipage}[ht]{0.23\textwidth}
\centering
\includegraphics[width=1.1\textwidth]{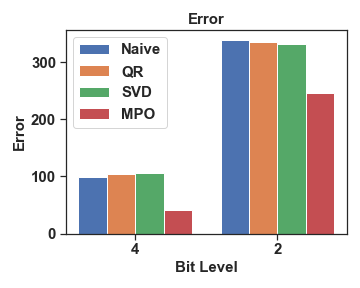} 
\end{minipage}
}
\caption{Comparison between MPO with other decomposition methods.}
\label{fig-decom}
\end{figure}
We compare the MPO decomposition in our approach with QR and SVD, which are popular decomposition methods. Results are in Figure~\ref{fig-decom}. 
Our method outperforms SVD and QR, with significantly lower quantization errors~(40.9 \vs 105.4 for SVD and 103.7 for QR at 4-bit precision) while introducing slight parameters. Additionally, MPO offers flexible tensor shapes, unlike QR and SVD which have fixed shapes, allowing us to balance accuracy and performance by adjusting quantization granularity.

\begin{figure}[t]
\centering
\subfigure[Memory Cost.]{
\label{fig-mem}
\begin{minipage}[ht]{0.23\textwidth}
\centering
\includegraphics[width=1.05\textwidth]{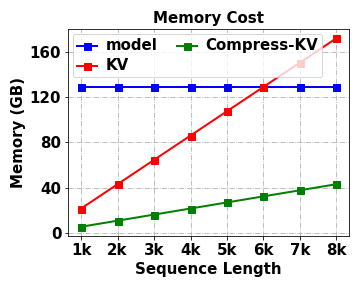} 
\end{minipage}
}
\subfigure[IO Latency.]{
\label{fig-latency}
\begin{minipage}[ht]{0.23\textwidth}
\centering
\includegraphics[width=1.05\textwidth]{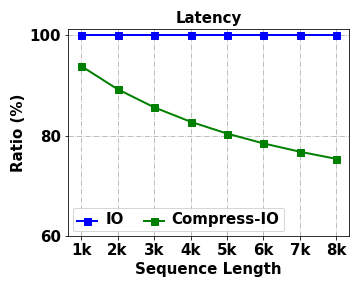} 
\end{minipage}
}
\caption{Efficiency of DecoQuant in terms of memory cost and IO latency.}
\label{fig-efficiency}
\end{figure}

\subsection{Analysis of the Efficiency}
\label{sec-analysis-complexity}
\paragraph{Memory and Latency.}
In this section, we provide additional analysis to show that memory and latency costs can be significantly reduced by our approach in the decoding phase.
Without loss of generality, we focus on LLaMA architecture~(70B), a popular open-source decoder-only model, and the sequence length of 1k to 8k for evaluation. 
In Figure~\ref{fig-mem}, we observe a significant reduction in the memory usage of the KV cache through compression, particularly evident when the sequence length reaches 6k. At this point, the cache size has matched the model size, while our cache remains under 30GB. Examining the latency in Figure~\ref{fig-latency}, we note that DecoQuant achieves even lower latency. These findings indicate that despite DecoQuant's increased computational effort, it remains negligible when compared to the communication overhead saved, ultimately resulting in latency optimization.
\section{Related Work}
In this section, we present related works in three aspects as well as draw distinctions of our approach to existing literature.

\paratitle{Tensor Decomposition for Language Models.}
Tensor decomposition~\cite{oseledets2011tensortrain} was first introduced to compress the neural network~\cite{novikov2015tensorizing}. Then it is used to achieve a better representation~\cite{gao2019compress}. In language modeling, tensor decomposition methods enable fine-grained model compression and tuning by decomposing the model's weights, and show a very high potential since such operations are independent of the model's structure. For example, in compression methods~\cite{gao2019compress,sun2020model}, in fine-tuning methods~\cite{gao2023small,liu2021enabling}, and in the field of pre-training~\cite{gao2022parameter,liu2023enhancing}. However the references of tensor decomposition in parameter quantization have not been well studied, and the contribution of this paper bridges the gap.

\paratitle{Quantization for LLMs.}
Quantization methods have been shown to be effective in reducing the size of the model as well as speeding it up. For example method~\cite{elias2022gptq} focuses on weight quantization while method~\cite{tim2022llmint8} focuses on activation value quantization. Activation value quantization is considered more challenging due to the presence of outliers. To address this issue, \citet{tim2022llmint8} cache the outlier values, while effective but still need to retain some of the FP16 values, thus making it difficult to achieve higher compression rates. A lot of quantization still needs to provide calibrated datasets, which may be difficult for some practical applications, \emph{e.g.}, users' private data are usually not allowed to be publicly accessible. This paper, on the other hand, addresses the activation value quantization methods still under the condition of no calibration.

\paratitle{KV Cache Compression.}
The decoding part of the current inference phase of LLM is mainly memory-bandwidth bound and an important approach is to alleviate the frequency of IO by compressing the KV cache. To achieve this goal, a straightforward approach is parameter quantization, but higher compression rates cannot be achieved due to the difficulty of activation value quantization. Another mainstream branch of research is concerned with reducing the number of tokens in the context, \emph{e.g.}, H2O~\cite{zhang2023h2o} by scores of attention. Other research is concerned with replacing the hard context with a soft prompt, \emph{e.g.}, AutoCompressors~\cite{chevalier2023adapting} by compressing the context into limited tokens. However, it may not be appropriate to choose to remove some tokens that are not important for the future only based on the existing context. Compared to the previous one, our approach keeps all tokens and ensures the integrity of the context.
\section{Conclusion}
In this paper, we proposed DecoQuant, a new data-free quantization method designed specifically for KV cache compression, to improve data generation efficiency.  
By first decomposing the KV cache matrices into local tensors, our approach only quantized the large local tensor with the major proportion of parameters in low-bit precision while maintained the small tensor in 16-bit precision. This approach can  mitigate the quantization difficulty from the original matrix to the small local tensor, which effectively reduces the quantization error in KV cache compression. During inference, we also developed an efficient dequantization technique based on 
the fused kernel tailored for dequantization of DecoQuant
to accelerate the generation process. Extensive experiments have demonstrated the effectiveness of the proposed approach in reducing the memory consumption of the KV cache and achieving competitive performance.
For future work,  we plan to explore the potential of leveraging Decoquant for scenarios where communication overhead plays a dominant role in LLM inference, specifically in the Splitwise technique where prefilling and decoding phases are in different nodes. 
\section{Limitations}
While we present promising results and contributions to the field, it is not without its limitations. The performance of our methods may be influenced by external factors such as hardware configurations, software dependencies, and environmental conditions. A thorough analysis of these factors and their impact on the performance of our methods is essential for practical deployment and real-world applications. In addition, our approach may facilitate the deployment of large language models onto a wide range of edge devices, including personal smartphones. However, this expansion may raise social concerns. It is crucial to consider potential biases and fairness issues in real-world applications.

\section{Acknowledgments}
This work was partially supported by National Natural Science Foundation of China under Grant No. 62222215 and 62206299, and Beijing Natural Science Foundation under Grant No. 4222027. Xin Zhao is the corresponding author.

\bibliography{acl_latex}

\begin{thebibliography}{27}
\expandafter\ifx\csname natexlab\endcsname\relax\def\natexlab#1{#1}\fi

\bibitem[{Chevalier et~al.(2023)Chevalier, Wettig, Ajith, and Chen}]{chevalier2023adapting}
Alexis Chevalier, Alexander Wettig, Anirudh Ajith, and Danqi Chen. 2023.
\newblock \href {https://aclanthology.org/2023.emnlp-main.232} {Adapting language models to compress contexts}.
\newblock In \emph{Proceedings of the 2023 Conference on Empirical Methods in Natural Language Processing, {EMNLP} 2023, Singapore, December 6-10, 2023}, pages 3829--3846. Association for Computational Linguistics.

\bibitem[{Dettmers et~al.(2022)Dettmers, Lewis, Belkada, and Zettlemoyer}]{tim2022llmint8}
Tim Dettmers, Mike Lewis, Younes Belkada, and Luke Zettlemoyer. 2022.
\newblock \href {https://doi.org/10.48550/ARXIV.2208.07339} {Llm.int8(): 8-bit matrix multiplication for transformers at scale}.
\newblock \emph{CoRR}, abs/2208.07339.

\bibitem[{Frantar et~al.(2022)Frantar, Ashkboos, Hoefler, and Alistarh}]{elias2022gptq}
Elias Frantar, Saleh Ashkboos, Torsten Hoefler, and Dan Alistarh. 2022.
\newblock \href {https://doi.org/10.48550/ARXIV.2210.17323} {{GPTQ:} accurate post-training quantization for generative pre-trained transformers}.
\newblock \emph{CoRR}, abs/2210.17323.

\bibitem[{Gao et~al.(2020)Gao, Cheng, He, Xie, Zhao, Lu, and Xiang}]{gao2019compress}
Ze-Feng Gao, Song Cheng, Rong-Qiang He, Zhi-Yuan Xie, Hui-Hai Zhao, Zhong-Yi Lu, and Tao Xiang. 2020.
\newblock Compressing deep neural networks by matrix product operators.
\newblock \emph{Physical Review Research}, 2(2):023300.

\bibitem[{Gao et~al.(2022)Gao, Liu, Zhao, Lu, and Wen}]{gao2022parameter}
Ze{-}Feng Gao, Peiyu Liu, Wayne~Xin Zhao, Zhong{-}Yi Lu, and Ji{-}Rong Wen. 2022.
\newblock \href {https://aclanthology.org/2022.coling-1.288} {Parameter-efficient mixture-of-experts architecture for pre-trained language models}.
\newblock In \emph{Proceedings of the 29th International Conference on Computational Linguistics, {COLING} 2022, Gyeongju, Republic of Korea, October 12-17, 2022}, pages 3263--3273. International Committee on Computational Linguistics.

\bibitem[{Gao et~al.(2023)Gao, Zhou, Liu, Zhao, and Wen}]{gao2023small}
Ze{-}Feng Gao, Kun Zhou, Peiyu Liu, Wayne~Xin Zhao, and Ji{-}Rong Wen. 2023.
\newblock \href {https://doi.org/10.18653/V1/2023.ACL-LONG.212} {Small pre-trained language models can be fine-tuned as large models via over-parameterization}.
\newblock In \emph{Proceedings of the 61st Annual Meeting of the Association for Computational Linguistics (Volume 1: Long Papers), {ACL} 2023, Toronto, Canada, July 9-14, 2023}, pages 3819--3834. Association for Computational Linguistics.

\bibitem[{Jacob et~al.(2018)Jacob, Kligys, Chen, Zhu, Tang, Howard, Adam, and Kalenichenko}]{jacob2018quantization}
Benoit Jacob, Skirmantas Kligys, Bo~Chen, Menglong Zhu, Matthew Tang, Andrew~G. Howard, Hartwig Adam, and Dmitry Kalenichenko. 2018.
\newblock \href {https://doi.org/10.1109/CVPR.2018.00286} {Quantization and training of neural networks for efficient integer-arithmetic-only inference}.
\newblock In \emph{2018 {IEEE} Conference on Computer Vision and Pattern Recognition, {CVPR} 2018, Salt Lake City, UT, USA, June 18-22, 2018}, pages 2704--2713. Computer Vision Foundation / {IEEE} Computer Society.

\bibitem[{Kolda and Bader(2009)}]{kolda2009tensor}
Tamara~G. Kolda and Brett~W. Bader. 2009.
\newblock \href {https://doi.org/10.1137/07070111X} {Tensor decompositions and applications}.
\newblock \emph{{SIAM} Rev.}, 51(3):455--500.

\bibitem[{Lin et~al.(2023)Lin, Tang, Tang, Yang, Dang, and Han}]{lin2023awq}
Ji~Lin, Jiaming Tang, Haotian Tang, Shang Yang, Xingyu Dang, and Song Han. 2023.
\newblock \href {https://doi.org/10.48550/ARXIV.2306.00978} {{AWQ:} activation-aware weight quantization for {LLM} compression and acceleration}.
\newblock \emph{CoRR}, abs/2306.00978.

\bibitem[{Liu et~al.(2023{\natexlab{a}})Liu, Gao, Chen, Zhao, and Wen}]{liu2023enhancing}
Peiyu Liu, Ze-Feng Gao, Yushuo Chen, Xin Zhao, and Ji-Rong Wen. 2023{\natexlab{a}}.
\newblock \href {https://doi.org/10.18653/v1/2023.findings-emnlp.920} {Enhancing scalability of pre-trained language models via efficient parameter sharing}.
\newblock In \emph{Findings of the Association for Computational Linguistics: EMNLP 2023}, pages 13771--13785, Singapore. Association for Computational Linguistics.

\bibitem[{Liu et~al.(2021)Liu, Gao, Zhao, Xie, Lu, and Wen}]{liu2021enabling}
Peiyu Liu, Ze{-}Feng Gao, Wayne~Xin Zhao, Zhi{-}Yuan Xie, Zhong{-}Yi Lu, and Ji{-}Rong Wen. 2021.
\newblock \href {https://doi.org/10.18653/V1/2021.ACL-LONG.418} {Enabling lightweight fine-tuning for pre-trained language model compression based on matrix product operators}.
\newblock In \emph{Proceedings of the 59th Annual Meeting of the Association for Computational Linguistics and the 11th International Joint Conference on Natural Language Processing, {ACL/IJCNLP} 2021, (Volume 1: Long Papers), Virtual Event, August 1-6, 2021}, pages 5388--5398. Association for Computational Linguistics.

\bibitem[{Liu et~al.(2023{\natexlab{b}})Liu, Desai, Liao, Wang, Xie, Xu, Kyrillidis, and Shrivastava}]{liu2023scissorhands}
Zichang Liu, Aditya Desai, Fangshuo Liao, Weitao Wang, Victor Xie, Zhaozhuo Xu, Anastasios Kyrillidis, and Anshumali Shrivastava. 2023{\natexlab{b}}.
\newblock \href {https://doi.org/10.48550/ARXIV.2305.17118} {Scissorhands: Exploiting the persistence of importance hypothesis for {LLM} {KV} cache compression at test time}.
\newblock \emph{CoRR}, abs/2305.17118.

\bibitem[{Mu et~al.(2023)Mu, Li, and Goodman}]{mu2023learning-gist}
Jesse Mu, Xiang~Lisa Li, and Noah~D. Goodman. 2023.
\newblock \href {https://doi.org/10.48550/ARXIV.2304.08467} {Learning to compress prompts with gist tokens}.
\newblock \emph{CoRR}, abs/2304.08467.

\bibitem[{Novikov et~al.(2015)Novikov, Podoprikhin, Osokin, and Vetrov}]{novikov2015tensorizing}
Alexander Novikov, Dmitry Podoprikhin, Anton Osokin, and Dmitry~P. Vetrov. 2015.
\newblock \href {https://proceedings.neurips.cc/paper/2015/hash/6855456e2fe46a9d49d3d3af4f57443d-Abstract.html} {Tensorizing neural networks}.
\newblock In \emph{Advances in Neural Information Processing Systems 28: Annual Conference on Neural Information Processing Systems 2015, December 7-12, 2015, Montreal, Quebec, Canada}, pages 442--450.

\bibitem[{Oseledets(2011)}]{oseledets2011tensortrain}
Ivan~V. Oseledets. 2011.
\newblock \href {https://doi.org/10.1137/090752286} {Tensor-train decomposition}.
\newblock \emph{{SIAM} J. Sci. Comput.}, 33(5):2295--2317.

\bibitem[{Paperno et~al.(2016)Paperno, Kruszewski, Lazaridou, Pham, Bernardi, Pezzelle, Baroni, Boleda, and Fern{\'{a}}ndez}]{paperno2016lambada}
Denis Paperno, Germ{\'{a}}n Kruszewski, Angeliki Lazaridou, Quan~Ngoc Pham, Raffaella Bernardi, Sandro Pezzelle, Marco Baroni, Gemma Boleda, and Raquel Fern{\'{a}}ndez. 2016.
\newblock \href {https://doi.org/10.18653/V1/P16-1144} {The {LAMBADA} dataset: Word prediction requiring a broad discourse context}.
\newblock In \emph{Proceedings of the 54th Annual Meeting of the Association for Computational Linguistics, {ACL} 2016, August 7-12, 2016, Berlin, Germany, Volume 1: Long Papers}. The Association for Computer Linguistics.

\bibitem[{Patel et~al.(2023)Patel, Choukse, Zhang, Goiri, Shah, Maleki, and Bianchini}]{patel2023splitwise}
Pratyush Patel, Esha Choukse, Chaojie Zhang, {\'{I}}{\~{n}}igo Goiri, Aashaka Shah, Saeed Maleki, and Ricardo Bianchini. 2023.
\newblock \href {https://doi.org/10.48550/ARXIV.2311.18677} {Splitwise: Efficient generative {LLM} inference using phase splitting}.
\newblock \emph{CoRR}, abs/2311.18677.

\bibitem[{Pope et~al.(2022)Pope, Douglas, Chowdhery, Devlin, Bradbury, Levskaya, Heek, Xiao, Agrawal, and Dean}]{pope2022efficiently}
Reiner Pope, Sholto Douglas, Aakanksha Chowdhery, Jacob Devlin, James Bradbury, Anselm Levskaya, Jonathan Heek, Kefan Xiao, Shivani Agrawal, and Jeff Dean. 2022.
\newblock \href {https://doi.org/10.48550/ARXIV.2211.05102} {Efficiently scaling transformer inference}.
\newblock \emph{CoRR}, abs/2211.05102.

\bibitem[{Rabanser et~al.(2017)Rabanser, Shchur, and G{\"{u}}nnemann}]{rabanser2017introduction}
Stephan Rabanser, Oleksandr Shchur, and Stephan G{\"{u}}nnemann. 2017.
\newblock \href {http://arxiv.org/abs/1711.10781} {Introduction to tensor decompositions and their applications in machine learning}.
\newblock \emph{CoRR}, abs/1711.10781.

\bibitem[{Sun et~al.(2020)Sun, Gao, Lu, Li, and Yan}]{sun2020model}
Xingwei Sun, Ze-Feng Gao, Zhong-Yi Lu, Junfeng Li, and Yonghong Yan. 2020.
\newblock A model compression method with matrix product operators for speech enhancement.
\newblock \emph{IEEE/ACM Transactions on Audio, Speech, and Language Processing}, 28:2837--2847.

\bibitem[{Touvron et~al.(2023)Touvron, Lavril, Izacard, Martinet, Lachaux, Lacroix, Rozi{\`{e}}re, Goyal, Hambro, Azhar, Rodriguez, Joulin, Grave, and Lample}]{touvron2023llama}
Hugo Touvron, Thibaut Lavril, Gautier Izacard, Xavier Martinet, Marie{-}Anne Lachaux, Timoth{\'{e}}e Lacroix, Baptiste Rozi{\`{e}}re, Naman Goyal, Eric Hambro, Faisal Azhar, Aur{\'{e}}lien Rodriguez, Armand Joulin, Edouard Grave, and Guillaume Lample. 2023.
\newblock \href {https://doi.org/10.48550/ARXIV.2302.13971} {Llama: Open and efficient foundation language models}.
\newblock \emph{CoRR}, abs/2302.13971.

\bibitem[{Wang et~al.(2010)Wang, Lin, and Yi}]{wang2010kernel}
Guibin Wang, Yisong Lin, and Wei Yi. 2010.
\newblock \href {https://doi.org/10.1109/GREENCOM-CPSCOM.2010.102} {Kernel fusion: An effective method for better power efficiency on multithreaded {GPU}}.
\newblock In \emph{2010 {IEEE/ACM} Int'l Conference on Green Computing and Communications, GreenCom 2010, {\&} Int'l Conference on Cyber, Physical and Social Computing, CPSCom 2010, Hangzhou, China, December 18-20, 2010}, pages 344--350. {IEEE} Computer Society.

\bibitem[{Xiao et~al.(2023)Xiao, Lin, Seznec, Wu, Demouth, and Han}]{xiao2023smoothquant}
Guangxuan Xiao, Ji~Lin, Micka{\"{e}}l Seznec, Hao Wu, Julien Demouth, and Song Han. 2023.
\newblock \href {https://proceedings.mlr.press/v202/xiao23c.html} {Smoothquant: Accurate and efficient post-training quantization for large language models}.
\newblock In \emph{International Conference on Machine Learning, {ICML} 2023, 23-29 July 2023, Honolulu, Hawaii, {USA}}, volume 202 of \emph{Proceedings of Machine Learning Research}, pages 38087--38099. {PMLR}.

\bibitem[{Zhang et~al.(2022)Zhang, Roller, Goyal, Artetxe, Chen, Chen, Dewan, Diab, Li, Lin, Mihaylov, Ott, Shleifer, Shuster, Simig, Koura, Sridhar, Wang, and Zettlemoyer}]{zhang2022opt}
Susan Zhang, Stephen Roller, Naman Goyal, Mikel Artetxe, Moya Chen, Shuohui Chen, Christopher Dewan, Mona~T. Diab, Xian Li, Xi~Victoria Lin, Todor Mihaylov, Myle Ott, Sam Shleifer, Kurt Shuster, Daniel Simig, Punit~Singh Koura, Anjali Sridhar, Tianlu Wang, and Luke Zettlemoyer. 2022.
\newblock \href {https://doi.org/10.48550/ARXIV.2205.01068} {{OPT:} open pre-trained transformer language models}.
\newblock \emph{CoRR}, abs/2205.01068.

\bibitem[{Zhang et~al.(2023)Zhang, Sheng, Zhou, Chen, Zheng, Cai, Song, Tian, R{\'{e}}, Barrett, Wang, and Chen}]{zhang2023h2o}
Zhenyu Zhang, Ying Sheng, Tianyi Zhou, Tianlong Chen, Lianmin Zheng, Ruisi Cai, Zhao Song, Yuandong Tian, Christopher R{\'{e}}, Clark~W. Barrett, Zhangyang Wang, and Beidi Chen. 2023.
\newblock \href {https://doi.org/10.48550/ARXIV.2306.14048} {H\({}_{\mbox{2}}\)o: Heavy-hitter oracle for efficient generative inference of large language models}.
\newblock \emph{CoRR}, abs/2306.14048.

\bibitem[{Zhao et~al.(2023)Zhao, Zhou, Li, Tang, Wang, Hou, Min, Zhang, Zhang, Dong, Du, Yang, Chen, Chen, Jiang, Ren, Li, Tang, Liu, Liu, Nie, and Wen}]{wayne2023survey}
Wayne~Xin Zhao, Kun Zhou, Junyi Li, Tianyi Tang, Xiaolei Wang, Yupeng Hou, Yingqian Min, Beichen Zhang, Junjie Zhang, Zican Dong, Yifan Du, Chen Yang, Yushuo Chen, Zhipeng Chen, Jinhao Jiang, Ruiyang Ren, Yifan Li, Xinyu Tang, Zikang Liu, Peiyu Liu, Jian{-}Yun Nie, and Ji{-}Rong Wen. 2023.
\newblock \href {https://doi.org/10.48550/ARXIV.2303.18223} {A survey of large language models}.
\newblock \emph{CoRR}, abs/2303.18223.

\bibitem[{Zhong et~al.(2024)Zhong, Liu, Chen, Hu, Zhu, Liu, Jin, and Zhang}]{zhong2024distserve}
Yinmin Zhong, Shengyu Liu, Junda Chen, Jianbo Hu, Yibo Zhu, Xuanzhe Liu, Xin Jin, and Hao Zhang. 2024.
\newblock \href {http://arxiv.org/abs/2401.09670} {Distserve: Disaggregating prefill and decoding for goodput-optimized large language model serving}.

\end{thebibliography}

\appendix
\section{Appendix}
\subsection{Analysis of Outliers}
\label{app-outlier}
The \emph{Interquartile Range}~(IQR) denotes the range between the 25th and 75th percentiles of the data. Outliers are often defined as data points that fall outside 1.5 times the IQR above the third quartile or below the first quartile.
Thus, to better understand the benefit of the distribution of outliers after MPO decomposition, we investigated the IQR in other layers~(1st, 16th, and 31st layers) and other structures~(keys and values).  

As seen in Table~\ref{tab-distribution_app}, we summarize the IQR of the target tensors. 
We observe, as discovered in Figure~\ref{fig-analysis}, that the IQR range of $\Tensor{T}_L$ is the narrowest, followed by $\Tensor{T}_S$, and the numerical ranges of the decomposed $\Tensor{T}_L$ and $\Tensor{T}_S$ are much smaller than those of the matrix. This indicates that our method can be universally applied to all key/value tensors.
\begin{table*}[t]
    \centering
    \small
    \caption{Analysis of the outlier distributions in LLaMA-7B.}
    \begin{tabular}{ccccccccc}
        \toprule
                                    &                            &                 & \multicolumn{3}{c}{Keys}  & \multicolumn{3}{c}{Values} \\
                                    &                            &                 & Q1   & Q3 & IQR    & Q1   & Q3 & IQR     \\ \midrule
        \multirow{9}{*}{LLaMA-7B}   &\multirow{3}{*}{1st layer}  & matrix          & -0.439&0.442&0.881&-0.013&0.013&0.026\\ 
                                    &                            & $\Tensor{T}_L$  & -0.027&0.027&0.055&-0.055&0.055&0.111\\ 
                                    &                            & $\Tensor{T}_S$  & -0.155&0.156&0.312&-0.228&0.222&0.449\\ \cmidrule{2-9}
                                    &\multirow{3}{*}{16th layer} & matrix          & -0.674&0.668&1.342&-0.307&0.306&0.613\\ 
                                    &                            & $\Tensor{T}_L$  & -0.056&0.056&0.112&-0.055&0.055&0.111 \\ 
                                    &                            & $\Tensor{T}_S$  & -0.238&0.232&0.470&-0.228&0.222&0.449\\ \cmidrule{2-9}
                                    &\multirow{3}{*}{32th layer} & matrix          & -0.685&0.672&1.357&-0.362&0.374&0.735\\ 
                                    &                            & $\Tensor{T}_L$  & -0.055&0.055&0.111&-0.055&0.055&0.111\\ 
                                    &                            & $\Tensor{T}_S$  & -0.228&0.222&0.449&-0.228&0.222&0.449\\ \bottomrule
    \end{tabular}
    \label{tab-distribution_app}
\end{table*}
\subsection{Details of the datasets}
In our in-context learning experiments, the length of the KV cache can be measured using the length of demonstrations since these demonstrations constitute the majority of the prefilling process. Therefore, we report the token per demonstration for five datasets to represent this, as shown in the Table~\ref{tab-app-icl}. We find that the datasets we used covered a range of context lengths, including longer contexts~(Boolq), shorter contexts~(Mr and Subj), and moderate contexts (Ag\_news and RTE).
\begin{table}[t!]
    \centering
    \begin{tabular}{ccc}
    \toprule
         \multirow{2}{*}{Dataset} & \multicolumn{2}{c}{Tokens per demonstration}  \\ 
                                  & OPT-based models & LLaMA-based models \\ \midrule
        Mr & 36 & 40\\
        Subj & 40 & 40\\ 
        Ag\_news & 65 & 75\\ 
        RTE & 75 & 85\\ 
        Boolq & 165 & 170\\ \bottomrule
    \end{tabular}
    \caption{Details of the datasets used for in-context learning. ``Tokens per demonstration'' indicates how long the demonstrations are for the average example.}
    \label{tab-app-icl}
\end{table}


\end{document}